\documentclass{article}
\usepackage{spconf,amsmath,graphicx}
\usepackage{mwe}
\usepackage{subcaption}
\usepackage{adjustbox}
\usepackage{float}
\usepackage{url}
\usepackage{times}
\usepackage{epsfig}
\usepackage{graphicx}
\usepackage{amsmath}
\usepackage{amssymb}
\usepackage{booktabs}
\usepackage{multirow}
\usepackage{bm}
\usepackage{adjustbox}
\usepackage{pifont}
\usepackage{diagbox}
\usepackage{xcolor}
\usepackage{amsmath,mathtools}
\usepackage{algorithm}
\usepackage{cite}
\usepackage{color}
\usepackage{cuted}
\usepackage{etoolbox}
\usepackage[numbers]{natbib}
\usepackage[linkcolor=red,anchorcolor=blue,urlcolor=blue,colorlinks=true]{hyperref}

\newcommand{\datasetname}{OpenAnimalTracks}
\newcommand{\datasetnameshort}{OAT}
\newcommand{\eg}{\textit{e.g.}}
\newcommand{\ie}{\textit{i.e.}}
\newcommand{\etal}{\textit{et al.}}
\title{OpenAnimalTracks: A Dataset for Animal Track Recognition}
\name{Risa Shinoda\textsuperscript{1,†}, Kaede Shiohara\textsuperscript{2,†}}
\address{\textsuperscript{1}Kyoto University, \textsuperscript{2}The University of Tokyo\\ 
{\small\textsuperscript{†}These authors contributed equally to this work.}}

\begin{document}

\maketitle

\begin{strip}\centering
\vspace{-0.3in}
\includegraphics[width=\textwidth]{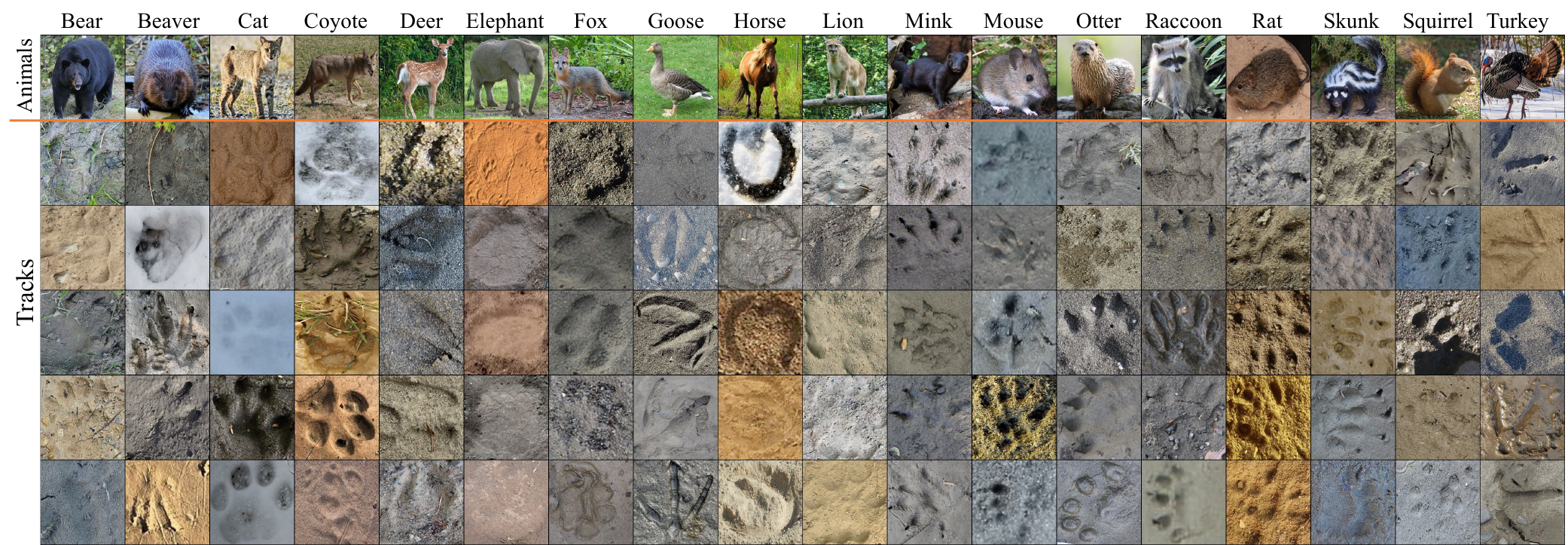}
 \vspace{-.1in}
\captionof{figure}{\textbf{OpenAnimalTracks dataset.} Our dataset consists of 3579 animal footprint images across 18 species under various environment and texture labels (mud, sand, and snow). This is the first publicly available animal footprint dataset.}

\label{fig:teaser}
\end{strip}

\begin{abstract}
Animal habitat surveys play a critical role in preserving the biodiversity of the land. One of the effective ways to gain insights into animal habitats involves identifying animal footprints, which offers valuable information about species distribution, abundance, and behavior.
However, due to the scarcity of animal footprint images, there are no well-maintained public datasets, preventing recent advanced techniques in computer vision from being applied to animal tracking.
In this paper, we introduce \datasetname~dataset, the first publicly available labeled dataset designed to facilitate the automated classification and detection of animal footprints. It contains various footprints from 18 wild animal species.
Moreover, we build benchmarks for species classification and detection and show the potential of automated footprint identification with representative classifiers and detection models. We find SwinTransformer achieves a promising classification result, reaching 69.41\% in terms of the averaged accuracy. Faster-RCNN achieves mAP of 0.295.
We hope our dataset paves the way for automated animal tracking techniques, enhancing our ability to protect and manage biodiversity.
Our dataset and code are available on GitHub\footnote{ \url{https://github.com/dahlian00/OpenAnimalTracks}}.
\end{abstract}
\section{Introduction}
\label{sec:intro}
Habitat surveys are essential to understanding and protecting the biodiversity of ecosystems, forming the foundation for effective conservation strategies and management practices. Obtaining precise information on the presence, abundance, and distribution of animal species contributes to preserving both individual species and their habitats.
A multitude of methods are employed to enhance our understanding of animal habitats, including camera traps, population surveys, and drones. 
However, it is often challenging to visually identify all the animals themselves.
To overcome this difficulty, animal tracking, which involves identifying species through their footprints, offers a complementary approach to studying animal habitats. Footprints can reveal valuable information about their species, behaviors, population, and movement patterns. Nonetheless, the process of discerning species based on footprints demands considerable expertise and experience.

In recent years, the rapid advancements in computer vision have grown research on applying computer vision techniques to the study of animals and their ecosystems.
In the field of animal observation, previous research conducted on species identification, behavior recognition~\cite{HOSSEININOORBIN2021106241, WANG2023107647,depthbehaviour}, monitoring system~\cite{ZHENG2023107857}, and camera trapping~\cite{Cunha_2021_CVPR, https://doi.org/10.1111/2041-210X.13880}. These technologies have the potential of automated animal observation, which often needs expert knowledge. 
Also, the development of animal datasets plays an essential role in enhancing these computer vision fields~\cite{aodha_song_cui_sun_shepard_adam_perona_belongie_2018,parkhi12a, Snapshot, 10.1007/978-3-030-01270-0_28,Cao_2019_ICCV,Ng_2022_CVPR, Chen_2023_CVPR}.  
Establishing datasets and benchmarks can contribute to building machine vision models to help understand animals.

There also exists prior research in the domain of animal footprint identification. 
For species classification from animal footprints, Kistner \etal~\cite{otter} have conducted a classification of three distinct otter species.
Furthermore, individual identification through footprints has been explored for various species, including amur tigers~\cite{ALIBHAI2023101947}, rhinoceros~\cite{rhinoceros}, tapirus~\cite{Tapirus}, giant panda~\cite{panda}, and cheetahs~\cite{Jewell2016SpottingCI}.  
However, these previous researches use landmarks of footprints for classification rather than using images directly. Furthermore, they provide only the landmark data, so the original image datasets are not accessible. 

In this paper, we introduce \datasetname~dataset, a publicly available resource for animal footprint data. This dataset comprises a diverse collection of footprint images from 18 animal species. 
To consider its applicability across various conditions, we include images from various environments such as mud, soil, and snow and also annotate this texture information. 
We collected reliable resources from experts and institutes in the field and additional footprint images from citizen scientists. 
Additionally, we have established the benchmark for animal footprint species classification and detection with five classifiers. We find attention-based model performs well on the dataset. In particular, SwinTransformer achieves the averaged accuracy of 69.41\%.
We believe that the \datasetname~dataset will bridge the gap between computer vision and animal tracking, fostering innovation in the field and addressing the challenges associated with traditional, time-consuming manual animal tracking techniques and the need for specialized expertise. By making this dataset openly available, we aim to stimulate further research, contributing to the enhanced understanding, protection, and management of the biodiversity of our ecosystems.

Our main contributions are as follows;
\begin{itemize}
\item We present \datasetname, which is the first publicly accessible dataset of animal footprints. The \datasetname~dataset comprises a collection of 3579 images captured from 18 different species.
\item We establish the benchmark of the \datasetname~dataset to classify and detect animal footprints. SwinTransformer achieves the best averaged accuracy of 69.41\% for classification, and Faster-RCNN achieves the best mAP of 0.295 for detection. Our results demonstrate the viability of employing automated animal tracking techniques.
\end{itemize}

\section{Related Work}
\label{sec:relatedwork}
Building the dataset and benchmarks is an important step in using computer vision in animal tracking and ecological surveys, as highlighted by various studies~\cite{Snapshot, 10.1007/978-3-030-01270-0_28, gagne2021florida, Ng_2022_CVPR, Chen_2023_CVPR,parkhi12a,KhoslaYaoJayadevaprakashFeiFei_FGVC2011}.
However, existing datasets are mainly aimed at monitoring the animals themselves, there is a lack of resources for tracking animals through their footprints, which is also an important factor in ecological surveys. 
Although there have been studies on animal footprints, the accessibility of these image datasets remains limited.

\noindent \textbf{Animal monitoring.} 
Recognizing and detecting wild animals is essential for comprehending ecosystems. With computer vision techniques, monitoring these animals involves image-based analysis. Large image datasets focusing on species classification have greatly advanced our ability to accurately recognize different animal species~\cite{aodha_song_cui_sun_shepard_adam_perona_belongie_2018, song2019selfie, parkhi12a,KhoslaYaoJayadevaprakashFeiFei_FGVC2011}. These datasets are essential for direct observation and monitoring of animals. 
Shifting to video-based monitoring, camera traps have emerged as a valuable tool, that enhances our understanding of ecology~\cite{Snapshot, 10.1007/978-3-030-01270-0_28, gagne2021florida}. Additionally, there are datasets developed for understanding animal behavior through video analysis~\cite{Ng_2022_CVPR, Chen_2023_CVPR}.
While these image datasets are pivotal for direct observation and monitoring of animals, an alternative approach for ecological monitoring involves tracking animals through their footprints.

\noindent \textbf{Animal Footprint Recognition.}
Animal footprints offer various information, such as species identification, population size, and animal behavior. For the individual identification, Jewell~\textit{et al.} used footprint landmarks of black rhinos and identified individuals~\cite{jewell_alibhai_law_2001}. They put thirteen landmarks manually and created measurements with customized software called FIT. With the same technique software FIT, cheetah footprints~\cite{Jewell2016SpottingCI}, white rhino footprints~\cite{whiterhino}, and tiger footprints~\cite{ALIBHAI2023101947} were used for identifying individuals. Manual annotations are still necessary for landmark extraction, and in the testing process, users have to upload the images for their systems and check the results individually. 
For species identification, Kistner~\textit{et al.} worked on classifying three otter species using footprints~\cite{otter}. They manually annotated 11 landmarks on each image and employed models for species identification. Similarly, the study on giant panda footprints~\cite{panda} also identified sex and age classes involved manual annotation with seven landmarks and used linear discriminant analysis. For the habitat survey, Moreira~\textit{et al.} analyzed tapir footprint landmarks and utilized pair-wise discriminant analysis to estimate the number of target animals~\cite{Tapirus}. Although these prior works expand the possibility of new animal tracking techniques from their footprints, accumulating multiple footprint landmarks information requires a lot of human annotations, and the testing phase could be more efficient. Moreover, the images they use are not publicly available.

\begin{table*}[t]
    \begin{adjustbox}{width=0.99\textwidth}
    \begin{tabular}{lcccccccccccccccccc|c} \toprule
       &Bear & Beaver & Cat & Coyote & Deer & Elephant & Fox & Goose & Horse & Lion & Mink & Mouse & Otter & Raccoon & Rat & Skunk & Squirrel & Turkey&Total\\ 
      \midrule
       \midrule
       \multicolumn{20}{l}{\textit{Classification}}\\
       \#Train  & 186 & 96 & 171 & 123 & 194 & 32 & 126 & 141 & 71 & 95 & 139 & 129 & 100 & 317 & 116 & 122 & 251 & 105 & 2514\\
       \#Val  &27 & 12 & 28 & 17 & 28 & 3 & 15 & 16 & 12 & 13 & 17 & 24 & 11 & 37 & 21 & 14 & 32 & 19 & 346\\
       \#Test  & 53 & 26 & 53 & 32 & 50 & 10 & 39 & 36 & 14 & 26 & 50 & 38 & 30 & 79 & 36 & 40 & 76 & 31 & 719\\
       Total & 266 & 134 & 252 & 172 & 272 & 45 & 180 & 193 & 97 & 134 & 206 & 191 & 141 & 433 & 173 & 176 & 359 & 155 & 3579 \\
       \midrule
       \midrule
       \multicolumn{20}{l}{\textit{Detection}}\\
        \#Train & 165 & 53 & 120 & 92 & 149 & 23 & 93 & 84 & 39 & 74 & 91 & 61 & 69 & 176 & 63 & 92 & 182 & 93 & 1719 \\
        \#Val & 23 & 7 & 17 & 13 & 21 & 3 & 13 & 12 & 5 & 10 & 13 & 8 & 10 & 24 & 9 & 13 & 26 & 13 & 240 \\
        \#Test & 48 & 17 & 35 & 27 & 44 & 7 & 28 & 24 & 12 & 22 & 27 & 19 & 21 & 52 & 19 & 27 & 53 & 28 & 510 \\
        Total & 236 & 77 & 172 & 132 & 214 & 33 & 134 & 120 & 56 & 106 & 131 & 88 & 100 & 252 & 91 & 132 & 261 & 134 & 2469 \\
      \bottomrule
    \end{tabular}
    \end{adjustbox}
  \caption{\textbf{The distribution of \datasetname.} \#Train, \#Val, and \#Test denote the numbers of images for training, validation, and testing, respectively.
  Note that the number of images for classification is equal to the number of annotated bounding boxes for the detection task. Some images contain multiple bounding boxes in an image. Therefore, images for classification are larger than that for detection. }
  \label{tb:dataset_info}
\end{table*}

\section{OpenAnimalTracks Dataset}
Our goal in this work is to explore ways to assist in the identification of animal footprints for ecological surveys.
To this end, we propose a novel dataset \datasetname, specifically designed for animal track identification. 
The overview of our dataset is given in Table~\ref{tb:dataset_info}.
We also annotate the background environment (\eg, mud, sand, and snow) of the footprints for more accurate footprint recognition.
\subsection{Collection.}
\label{sec:data_collection}
Animal footprint identification needs expertise. Therefore, we collect reliable resources from experts and institutes in the field.

In addition, we gathered additional footprint images from the internet to boost the size of our dataset. To ensure the annotation quality, we carefully verified the collected images by confirming the footprints against their shapes and characteristic features.
These images encompass 18 distinct species: bear~(black bear), beaver~(american beaver), cat~(bobcat), coyote, deer~(mule deer), elephant~(asian and african elephant), fox~(gray fox), goose~(canada goose),  horse~(domestic horse), lion~(mountain lion), mouse~(western harvest mouse), otter~(river ottter), raccoon, rat~(california kangaroo rat, black rat), skunk~(stripped skunk, western spotted skunk), and squirrel~(western gray squirrel), and turkey~(wild turkey). 

\subsection{Annotations}
\noindent \textbf{Bounding box.} 
We annotated bounding boxes using LabelMe~\cite{labelme}.
We then cropped these bounding boxes to create a classification dataset and generated ground truth for a detection dataset.

\noindent \textbf{Data split.} We divide the raw images into train, validation, and test sets with a 7:1:2 ratio for each class.
The numbers of training/validation/testing images for classification and detection are reported in Table~\ref{tb:dataset_info}.
 
\begin{figure}
  \centering
  \begin{adjustbox}{width=1.0\linewidth}
   \includegraphics[width=\textwidth]{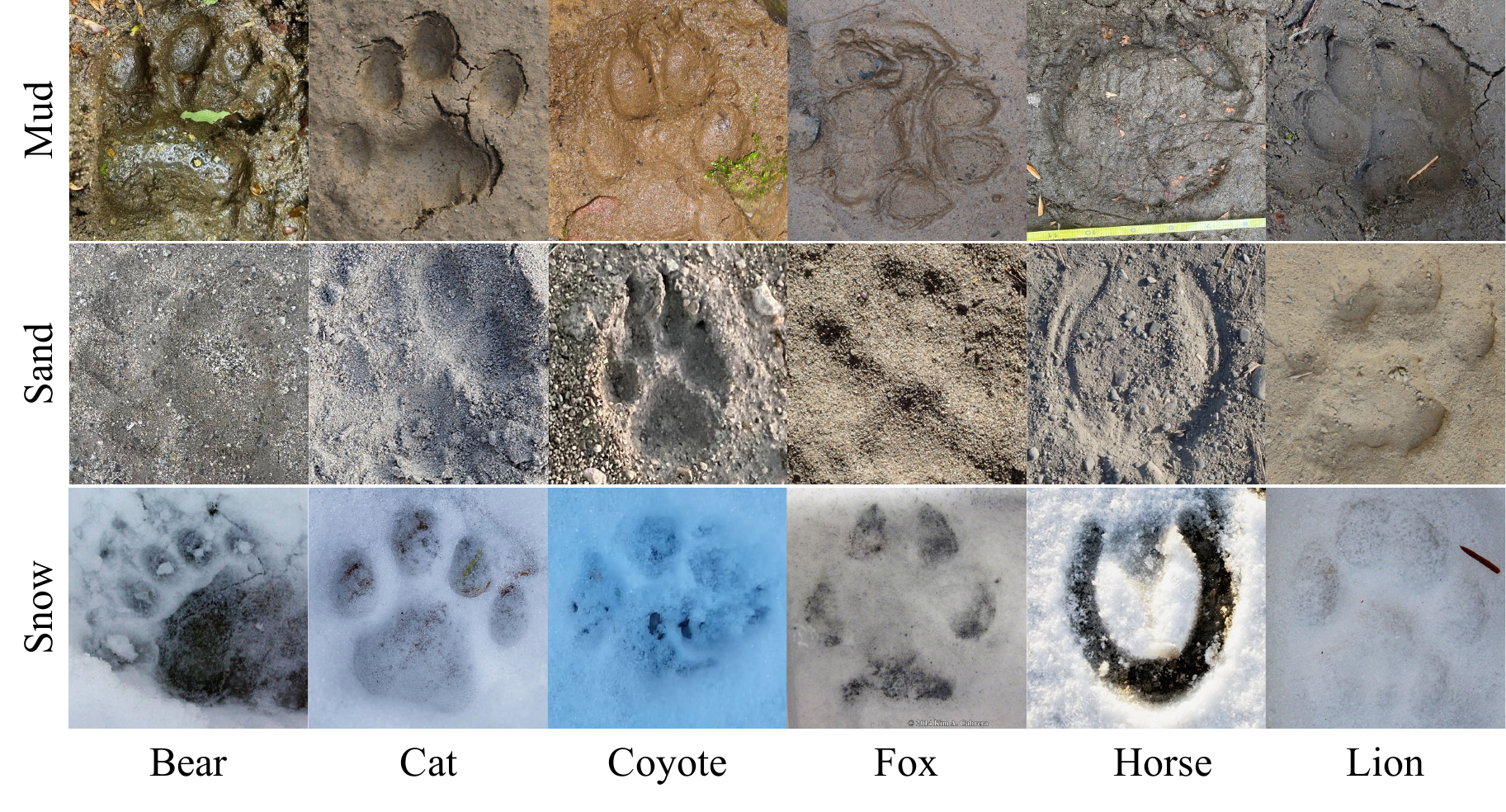}
   \end{adjustbox}
    \caption{\textbf{Texture annotations.} We divide the background texture into mud, sand, and snow.
}
  \label{fig:texture_example}
\end{figure}

\begin{figure}
  \centering
  \begin{adjustbox}{width=1.0\linewidth}
   \includegraphics[width=\textwidth]{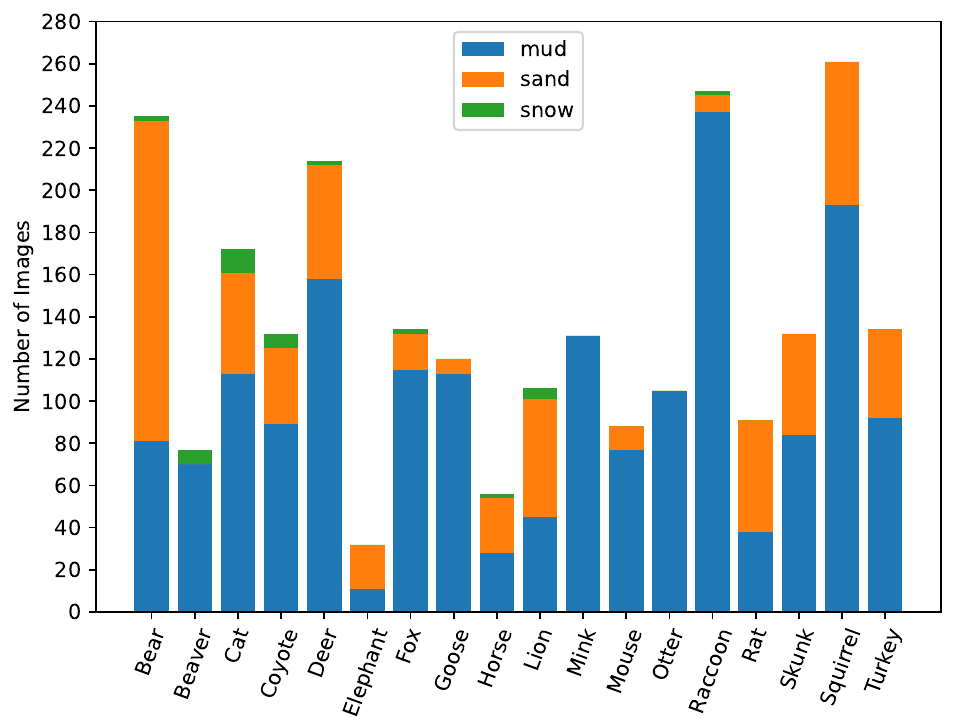}
   \end{adjustbox}
    \caption{\textbf{Texture distribution.} The animal footprints of our dataset are distributed in mud, sand, and snow.
}
  \label{fig:texture_distribution}
\end{figure}

\begin{table*} 
    \centering
    \begin{adjustbox}{width=0.99\textwidth}
    \begin{tabular}{lcccccccccccccccccc|c} \toprule
      \multirow{2}{*}{Model} &\multicolumn{19}{c}{Top-1 Accuracy (\%)}\\
      \cmidrule(lr){2-20}
      &Bear & Beaver & Cat & Coyote & Deer & Elephant & Fox & Goose & Horse & Lion & Mink & Mouse & Otter & Raccoon & Rat & Skunk & Squirrel & Turkey&Mean\\ 
      \midrule
      \midrule
      \multicolumn{20}{l}{\textit{Full tuning}}\\
       VGG-16  & 73.58 & 53.85 & 52.83 & 56.25 & 77.78 & 78.57 & 64.10 & 73.08 & 60.00 & 42.11 & 74.00 & 50.00 & 80.00 & 58.23 & 47.22 & 80.00 & 53.95 & 77.42 & 64.05\\
       Res-50  & 66.04 & 50.00 & 52.83 & 46.88 & 77.78 & 71.43 & 74.36 & 73.08 & 68.00 & 52.63 & 78.00 & 50.00 & 70.00 & 49.37 & 61.11 & 55.00 & 53.95 & 80.65 & 62.84\\
       Eff-b1  & 54.72 & 42.31 & 41.51 & 28.12 & 69.44 & 50.00 & 46.15 & 61.54 & 64.00 & 52.63 & 66.00 & 40.00 & 46.67 & 34.18 & 47.22 & 40.00 & 39.47 & 45.16 & 48.2\\
       ViT-B& 69.81 & 53.85 & 62.26 & 62.50 & 88.89 & 78.57 & 71.79 & 73.08 & 54.00 & 65.79 & 80.00 & 80.00 & 63.33 & 62.03 & 58.33 & 70.00 & 52.63 & 77.42 & 68.02\\
       Swin-B & 60.38 & 57.69 & 60.38 & 78.12 & 80.56 & 64.29 & 79.49 & 76.92 & 72.00 & 73.68 & 86.00 & 50.00 & 76.67 & 58.23 & 69.44 & 65.00 & 56.58 & 83.87 & \textbf{69.41}\\
      
      \midrule
      \midrule
      \multicolumn{20}{l}{\textit{Linear probing}}\\
       VGG-16  & 47.17 & 23.08 & 22.64 & 12.50 & 63.89 & 35.71 & 41.03 & 61.54 & 64.00 & 52.63 & 48.00 & 10.00 & 56.67 & 18.99 & 33.33 & 52.50 & 25.00 & 51.61 & 40.02\\
       Res-50  &  26.42 & 26.92 & 26.42 & 25.00 & 61.11 & 21.43 & 12.82 & 61.54 & 64.00 & 65.79 & 56.00 & 0.00 & 60.00 & 16.46 & 30.56 & 25.00 & 25.00 & 32.26 & 35.37\\
       Eff-b1  & 39.62 & 3.85 & 11.32 & 0.00 & 8.33 & 14.29 & 2.56 & 23.08 & 74.00 & 28.95 & 30.00 & 0.00 & 16.67 & 13.92 & 5.56 & 22.50 & 18.42 & 6.45 & 17.75\\
       ViT-B  &  47.17 & 19.23 & 35.85 & 25.00 & 72.22 & 42.86 & 43.59 & 65.38 & 64.00 & 57.89 & 72.00 & 30.00 & 66.67 & 24.05 & 36.11 & 37.50 & 30.26 & 51.61 & \textbf{45.63}\\
       Swin-B  & 18.87 & 23.08 & 39.62 & 18.75 & 52.78 & 50.00 & 41.03 & 50.00 & 70.00 & 84.21 & 62.00 & 0.00 & 43.33 & 17.72 & 41.67 & 25.00 & 19.74 & 45.16 & 39.05\\
      \bottomrule
    \end{tabular}
    \end{adjustbox}
  \caption{\textbf{The classification results on \datasetname by top-1 accuracy.} Attention-based models such as ViT-B and Swin-B outperform convolution-based models both on fine-tuning and linear probing. For all models, linear probing leads to significant drops in classification performance because of the lack of beneficial features of pretrained models on ImageNet. The best mean accuracy is in \textbf{bold}.}
  \label{tb:classification}
  \vspace{-10pt}
\end{table*}

\noindent \textbf{Texture.}
In the open-world scenario, animal footprints are often encountered in a variety of environmental contexts. 
Specifically, we observe that the images of our dataset include three types of background textures: mud, sand, and snow. 
The representative examples of the textures are shown in Fig.~\ref{fig:texture_example}.
We carefully categorize all of our images into the three texture categories.
As a result, our dataset consists mainly of footprints on mud and sand at $72\%$ and $26\%$, respectively.
We visualize the texture distributions in Fig.~\ref{fig:texture_distribution}.
And eight species, \ie, bear, beaver, cat, coyote, deer, fox, horse, lion, and raccoon, include footprints on snow, which is $2$\% of all images.
This diverse set of backgrounds aims to enhance the robustness and generality of models trained on our dataset.

\section{Footprint Classification}
\label{sec:experiment}
Here, we show the applicability of \datasetname~(OAT) dataset by building benchmarks for animal track classification on 18 species with state-of-the-art image classification baselines.

\subsection{Setup}
\noindent \textbf{Models.} We adopt five representative classifiers based on convolutional networks, \ie VGG-16~\cite{vgg}, ResNet-50~\cite{resnet} (Res-50), and EfficientNet-b1~\cite{efficientnet} (Eff-b1), and based on transformers~\cite{attention}, \ie, Vision Transformer~\cite{vit} (ViT-B) and SwinTransformer~\cite{swin} (Swin-B).
All models are pre-trained on ImageNet~\cite{imagenet}.

\vskip.3\baselineskip
\noindent \textbf{Preprocessing.} 
Input images are resized to $224^2$ pixels for VGG-16, Res-50, ViT-B, and Swin-B, and to $240^2$ pixels for Eff-b1. 
During training, we randomly change the brightness, contrast, saturation, hue, and compression rate of images, flip images vertically and horizontally, and rotate images to augment training samples. We adopt Albumentations~\cite{albumentations} to implement the augmentations.

\vskip.3\baselineskip
\noindent \textbf{Training.} 
We update model parameters based on the cross entropy loss. 
The models are trained for 100 epochs with batch size of 128, which is enough for the training losses of models to converge.
We use SGD optimizer and the learning rate is set to $1e{-4}$ for all the models.
Each training is conducted using a single NVIDIA A100 GPU.
We also adopt logit adjustment~\cite{logit_adjustment} to redress the class-imbalance of the training samples.

We conduct two types of training: \textit{full tuning} and \textit{linear probing}. 
In the case of full tuning, we update all parameters from initial weights pre-trained on ImageNet. 
On the contrary, in linear probing, we only update the weights of the last linear layer, and other weights are frozen during training, which makes it possible to measure the compatibility of pre-trained models on ImageNet for animal species classification from their footprints.

\vskip.3\baselineskip
\noindent \textbf{Metrics.} 
We adopt the class-wise top-1 accuracy and averaged top-1 accuracy over the classes to evaluate the classifiers. 

\subsection{Result}
\noindent \textbf{Full tuning.} 
We report the full tuning result on \datasetname~dataset in the top side of Table.~\ref{tb:classification}.
We can see that the attention-based methods (ViT-B and Swin-B) outperform the convolution-based methods (VGG-16, Res-50, and Eff-b1). 
This result implies attentions work better than convolutions on animal tracks where the structures are more important than the textures because attention-based models tend to focus on the structures of images rather than textures, as reported in ~\cite{zhang2022delving}.
The classifiers recognize deers well ($81\%$ for Swin) while they struggle with identifying mouses ($50\%$ for Swin). 

\vskip.3\baselineskip
\noindent \textbf{Linear probing.} 
We next report the linear probing result in the bottom side of Table.~\ref{tb:classification}.
We observe a similar tendency to the result of fine-tuning; attention-based models perform better than convolution-based models. 
However, the averaged accuracy significantly drops on linear probing compared to fine-tuning (\eg, from 68.02\% to 45.63\% for ViT-B), indicating that there is a large domain gap between animal track classification and general image classification such as ImageNet; therefore, there is a room for the improvements by exploring the specific approaches for animal species identification from their footprints.
\begin{figure}
  \centering
  \begin{adjustbox}{width=0.95\linewidth}
   \includegraphics[width=\textwidth]{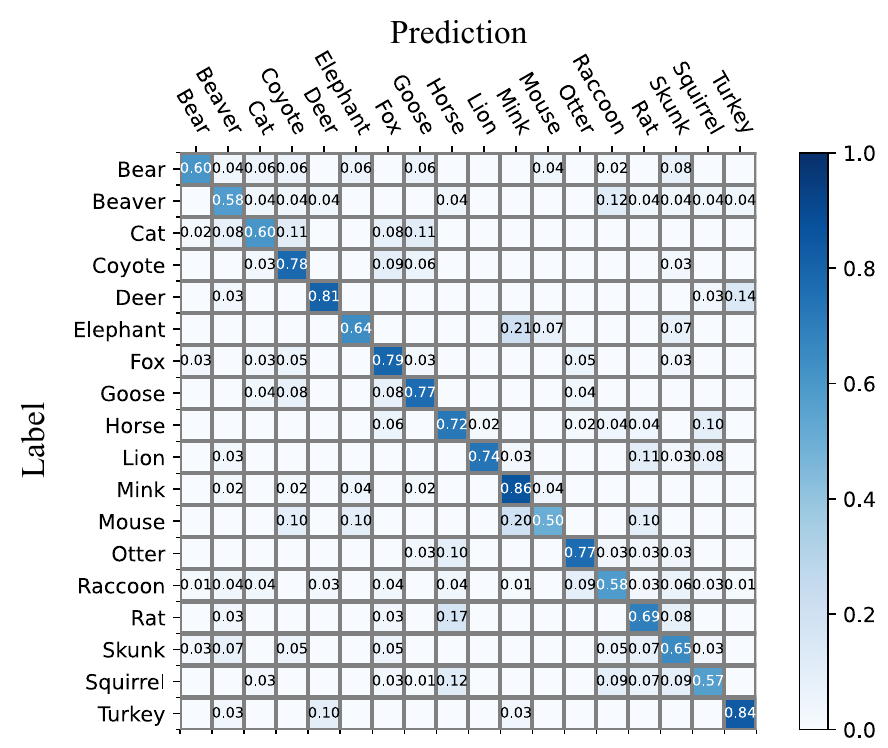}
   \end{adjustbox}
    \caption{\textbf{Confusion matrix of Swin-B.} Empty squares represent the ratio of 0.0. }
     \vspace{-10pt}
  \label{fig:confusion}
\end{figure}

\subsection{Analysis}
\noindent \textbf{Confusion matrix.} To investigate the prediction tendency of deep neural networks on \datasetname, we visualize the confusion matrix of Swin-B in Fig.~\ref{fig:confusion}.
We observed that a frequent misclassification occurs with coyotes being mistaken as foxes ($9\%$ for Swin-B). 
This is because footprint shapes between the two species 
are similar to each other; they are genetically close to each other.
Similarly, it can be seen that mouse footprints tend to be misclassified as minks ($20\%$ for Swin-B).
These failure cases can be improved by incorporating the size information of footprints into classifiers, which is left as future work.

\begin{figure}
  \centering
  \begin{adjustbox}{width=0.99\linewidth}
   \includegraphics[width=\textwidth]{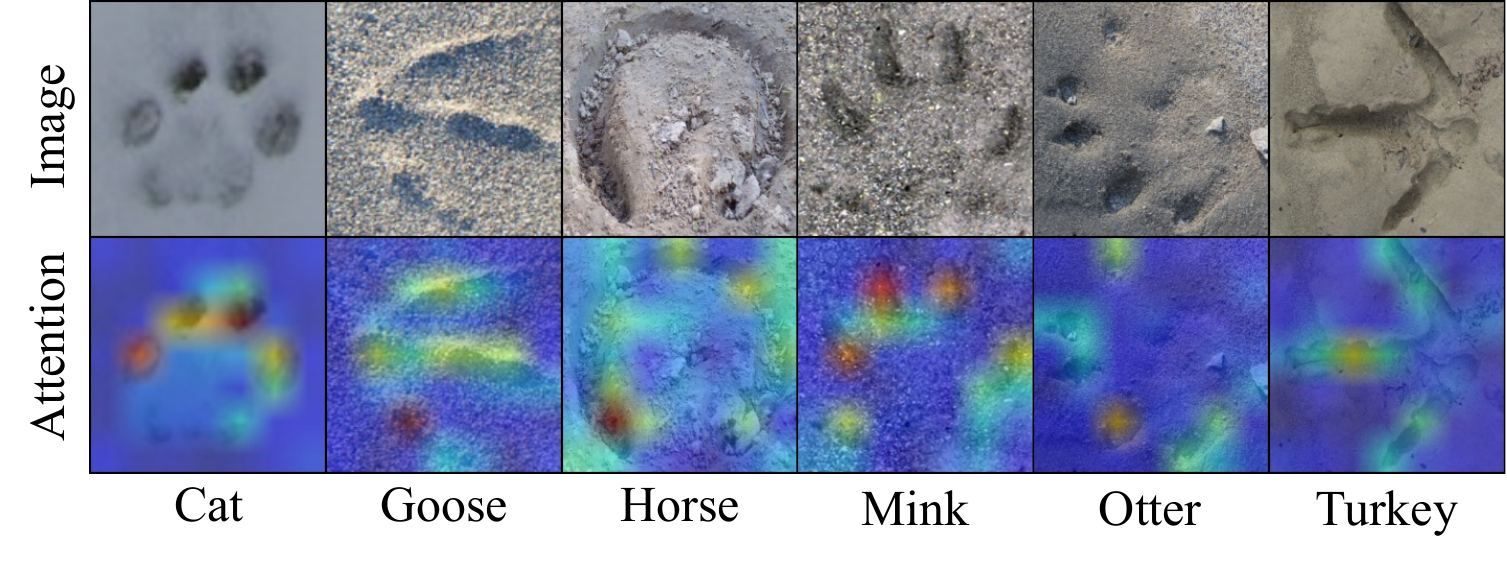}
   \end{adjustbox}
    \caption{\textbf{Attention map visualization of ViT-B.} The model properly pays attention to representative points of footprints. When the value is small, it is colorized in blue, when the value is large, it becomes red, and when the value is moderate, it passes through colors from green to yellow. }
  \label{fig:attention}
\end{figure}

\vskip.3\baselineskip
\noindent \textbf{Attention map.}
We visualize attention maps of ViT-B on successful cases, where the model predicts correctly the ground truth labels. 
For each sample, we compute the averaged attention score over all self-attention layers of ViT-B, and we normalize them into $[0, 1]$, excluding the class token. 
The result is given in Fig.~\ref{fig:attention}.
We find ViT accurately captures important landmarks (\eg, paw) in most cases.
This indicates that ViT predicts species based on the shape of tracks without shortcut cheating, \eg, focusing on class-specific environment and texture, which supports the high quality of our dataset.

\section{Footprint Detection}
For more practical case of animal tracking, we further investigate to detect footprints from raw (un-cropped) images using object detection models on our \datasetnameshort~dataset.

\subsection{Setup}
\noindent \textbf{Model.}
We adopt three conventional object detection models including Faster R-CNN~\cite{fasterrcnn}, SSD~\cite{ssd}, and YOLOv3~\cite{yolo}.
We use MMDetection~\cite{mmdetection} to arrange the models.
The preprocessing varies widely depending on the model.
We follow the original configuration of MMDetection~\cite{mmdetection} for the models.
Please refer to its implementations for more details.

\noindent \textbf{Training.}
Similarly to preprocessing, training strategies are different from models to models.
We mainly follow the original training methods,\eg, loss functions, optimizers, and learning rates while we train the models for 24 epochs and set the batch size to 8 for fair comparisons.

\noindent \textbf{Metrics.}
Following the convention of object detection, we adopt the average precision (AP) to evaluate models. 
AP represents the area under the Precision-Recall curve
Each AP is computed using an interpolation method at a set of eleven equally spaced recall levels. The mean AP (mAP) is computed by averaging AP over all classes.
To further refine the evaluation, we also consider mAP at specific Intersection-over-Union (IoU) thresholds, commonly set at 0.5 and 0.75.
These metrics, represented as \textit{mAP$_{50}$} and \textit{mAP$_{75}$}, respectively.

\begin{table*} 
    \centering
    \begin{adjustbox}{width=0.99\textwidth}
    \begin{tabular}{lcccccccccccccccccc|c} \toprule
    \multirow{2}{*}{Model} &\multicolumn{19}{c}{Average Precision}\\
      \cmidrule(lr){2-20}
       &Bear & Beaver & Cat & Coyote & Deer & Elephant & Fox & Goose & Horse & Lion & Mink & Mouse & Otter & Raccoon & Rat & Skunk & Squirrel & Turkey&Mean\\ 
      \midrule
       Faster R-CNN  & 0.352 & 0.107 & 0.405 & 0.419 & 0.219 & 0.027 & 0.426 & 0.477 & 0.145 & 0.51 & 0.215 & 0.165 & 0.429 & 0.299 & 0.27 & 0.304 & 0.191 & 0.344 & \textbf{0.295}\\
       SSD  &  0.275 & 0.093 & 0.234 & 0.318 & 0.259 & 0.092 & 0.299 & 0.403 & 0.181 & 0.294 & 0.217 & 0.13 & 0.329 & 0.217 & 0.092 & 0.256 & 0.111 & 0.275 & 0.226\\
       YOLOv3  & 0.247 & 0.138 & 0.365 & 0.381 & 0.17 & 0.13 & 0.404 & 0.52 & 0.219 & 0.483 & 0.225 & 0.262 & 0.333 & 0.311 & 0.153 & 0.28 & 0.147 & 0.356 & 0.285\\
      \bottomrule
    \end{tabular}
    \end{adjustbox}
  \caption{\textbf{The detection results on \datasetname~by average precision.} Faster R-CNN achieves the best mean Average Precision. The best averaged accuracy is in \textbf{bold}.}
  \label{tb:detection_class}
\end{table*}

\begin{table}[t]
    \centering
    \begin{tabular}{lccc} \toprule
      Model &mAP & mAP$_{50}$& mAP$_{75}$\\ 
      \midrule
       Faster R-CNN  & \textbf{0.295} &0.494&\textbf{0.312} \\
       SSD  &  0.226 &0.447 &0.197 \\
       YOLOv3& 0.285 & \textbf{0.512} & 0.275 \\
      \bottomrule
    \end{tabular}
  \caption{\textbf{The detection results of different IoU 
 thresholds of mAP.} Faster R-CNN outperforms SSD and YOLOv3 in mAP and mAP$_{75}$ while YOLOv3 achieves the best mAP$_{50}$.}
  \label{tb:map_threshold}
  \vspace{-10pt}
\end{table}

\subsection{Result}
In the object detection task, we present class-wise results in Table~\ref{tb:detection_class}. We can see that the de-facto standard model, Faster R-CNN still works well on our dataset, had the highest mAP. It also had the most number of categories where it ranked first in performance. We observed that certain categories, such as goose and lion, consistently registered higher AP values. One possible explanation for this could be the distinctiveness of their features, making them easier for the model to recognize. Conversely, classes like beaver and elephant had lower AP values.
To improve performance on such underrepresented classes, future work could focus on techniques for class imbalance correction or the incorporation of more robust data augmentation methods.
We also evaluate the models on \datasetnameshort ~in terms of mAP, mAP$_{50}$, and mAP$_{75}$. 
We give the result in Table~\ref{tb:map_threshold}.
Faster R-CNN achieves the best mAP (0.295) and mAP$_{75}$ (0.312). 
YOLOv3 outperforms Faster R-CNN and SSD in mAP$_{50}$ but is inferior to Faster R-CNN
in mAP$_{75}$ (0.275 vs. 0.312). 

\begin{figure}
  \centering
  \begin{adjustbox}{width=0.48\textwidth}
   \includegraphics[width=\textwidth]{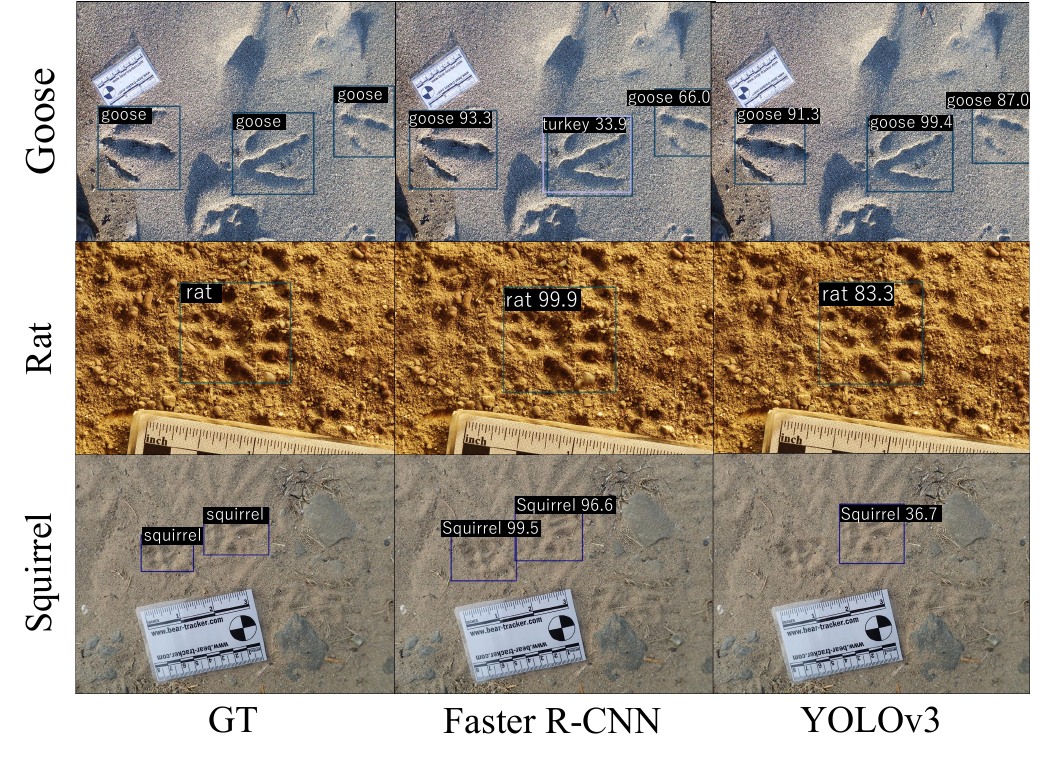}
   \end{adjustbox}
    \caption{\textbf{Visual detection results of Faster R-CNN and YOLOv3.} }
  \label{fig:detection}
  \vspace{-10pt}
\end{figure}

We visualize the detection results in Fig.~\ref{fig:detection}. We can see that models occasionally misidentify one species for another or fail to recognize footprints. Nonetheless, the models are capable of detecting footprints even when they are not deeply imprinted, as is the case with bear footprints.

\section{Conclusion \& Future Work}
We present \datasetname, the first open animal footprint image dataset for animal species identification.
We collect 2469 images across 18 species under various environments and carefully annotate 3579 bounding boxes for animal track classification and detection.
In addition, we establish benchmarks with five and three representative deep neural networks for classification and detection, respectively. 
The experimental result on classification indicates that attention-based models such as Vision Transformer perform well compared to convolution models on classification because structures are more important than textures to identify footprints.

For future work, we will enlarge our dataset in terms of the number of images per species, types of annotation (\eg, segmentation masks), and the range of species.
Also, we will explore the more effective methods/models to recognize animal tracks.

\section{Acknowledgements}\label{acknowledgements}
We sincerely thank Kim A. Cabrera from Beartracker’s Animal Tracks~\cite{beartracker}, the Wildlife Research Center of Kyoto University~\cite{wrc}, and Japan Wildlife Center~\cite{jwc} for their contribution to the images that enabled this research. 
We greatly thank Prof. Toshihiko Yamasaki from the University of Tokyo for providing computation resources.

\begingroup
    \setlength{\bibsep}{2pt}
\bibliographystyle{IEEE}
\bibliography{IEEE}
\endgroup

\end{document}